\documentclass{article}
\PassOptionsToPackage{dvipsnames}{xcolor}
\usepackage{iclr2025_conference,times}

\iclrfinalcopy 

\usepackage{amsmath,amsfonts,bm}

\def\eqref#1{equation~\ref{#1}}

\def\1{\bm{1}}

\def\va{{\bm{a}}}

\def\vc{{\bm{c}}}

\def\vx{{\bm{x}}}
\def\vy{{\bm{y}}}

\DeclareMathAlphabet{\mathsfit}{\encodingdefault}{\sfdefault}{m}{sl}
\SetMathAlphabet{\mathsfit}{bold}{\encodingdefault}{\sfdefault}{bx}{n}

\def\gD{{\mathcal{D}}}

\def\gF{{\mathcal{F}}}

\def\sR{{\mathbb{R}}}

\DeclareMathOperator*{\argmin}{arg\,min}

\usepackage{amssymb}
\usepackage{enumitem}
\usepackage{xspace}
\usepackage{booktabs}
\usepackage{multirow}
\usepackage{colortbl}
\usepackage{pifont}
\usepackage{graphicx}
\usepackage{siunitx}
\usepackage{amsmath}
\usepackage{makecell}
\usepackage{microtype}

\usepackage{caption}
\usepackage{algorithm}
\usepackage{algpseudocode}

\usepackage{url}

\def\modelname{LangToMo\xspace}
\def\modelshort{LTM\xspace}

\definecolor{Gray}{gray}{0.90}
\definecolor{lightblue}{rgb}{0.9,0.96,1}

\newcommand{\cmark}{\ding{51}}
\newcommand{\xmark}{\ding{55}}

\newcommand{\bhdr}[1]{\noindent\textbf{#1}}

\definecolor{iccvblue}{rgb}{0.21,0.49,0.74}
\usepackage[pagebackref,breaklinks,colorlinks,allcolors=iccvblue]{hyperref}
\usepackage{cleveref}

\title{Pixel Motion as Universal Representation \\for Robot Control}

\author{%
Kanchana Ranasinghe, 
Xiang Li, 
E-Ro Nguyen,
Cristina Mata, 
Jongwoo Park, \vspace{0.1em} \\
\textbf{Michael S Ryoo} \vspace{0.5em} \\
Stony Brook University \vspace{0.5em} \\
\texttt{kranasinghe@cs.stonybrook.edu} \\
}

\begin{document}
\maketitle

\vspace{-1.0em}
\begin{figure}[h]
    \centering
    \includegraphics[width=0.82\linewidth]{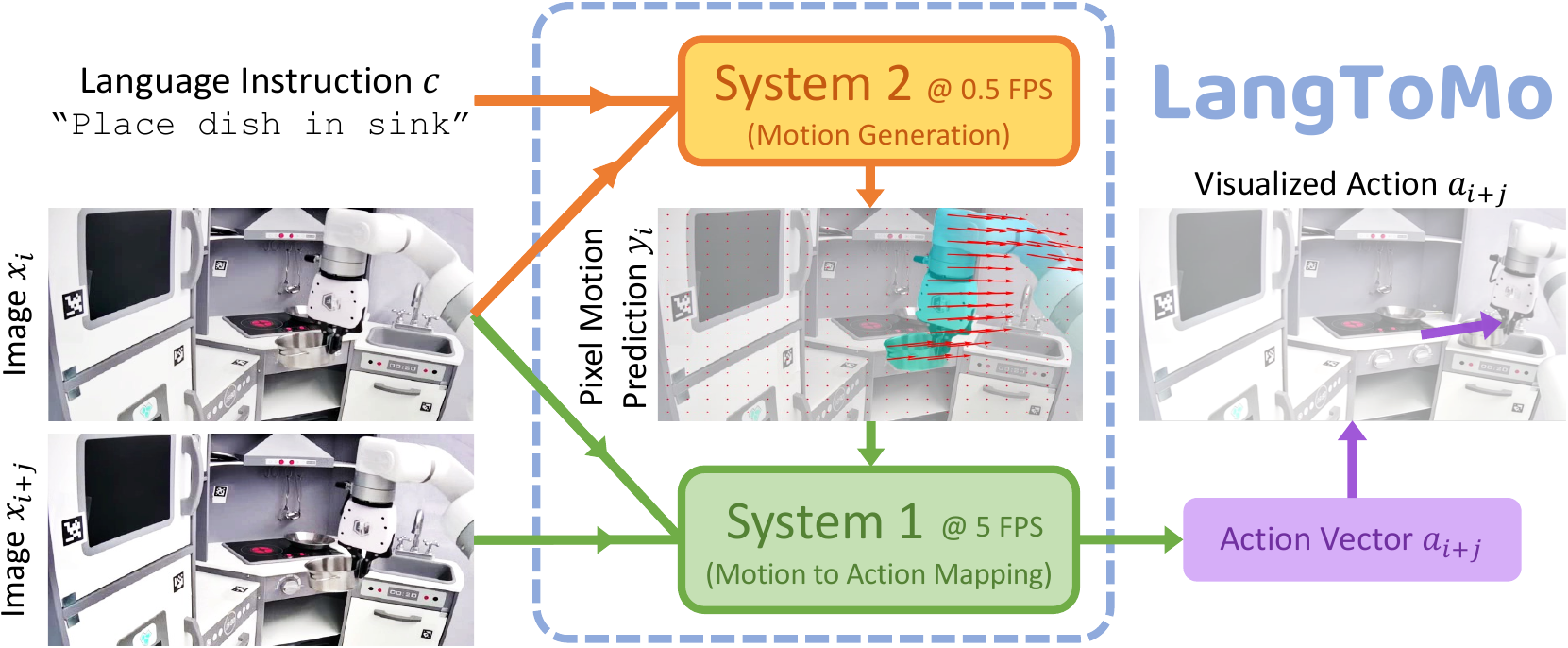}
    \vspace{-0.4em}
    \caption{
    Dual-System VLA Framework, \modelname, with pixel motion representations.
    }
    \label{fig:teaser}
\end{figure}

\begin{abstract}
    We present \modelname, a vision-language-action framework structured as a dual-system architecture that uses pixel motion forecasts as intermediate representations. 
    Our high-level \textit{System 2}, an image diffusion model, generates text-conditioned pixel motion sequences from a single frame to guide robot control.
    Pixel motion—a universal, interpretable, and motion-centric representation—can be extracted from videos in a weakly-supervised manner, enabling diffusion model training on any video-caption data.
    Treating generated pixel motion as learned \textit{universal representations}, our low level \textit{System 1} module translates these into robot actions via motion-to-action mapping functions, which can be either hand-crafted or learned with minimal supervision.
    System 2 operates as a high-level policy applied at sparse temporal intervals, while System 1 acts as a low-level policy at dense temporal intervals.
    This hierarchical decoupling enables flexible, scalable, and generalizable robot control under both unsupervised and supervised settings, bridging the gap between language, motion, and action.
    Checkout 
    \href{https://kahnchana.github.io/LangToMo}{\texttt{kahnchana.github.io/LangToMo}}
\end{abstract}


\section{Introduction}
\label{sec:intro}

Translating open-ended natural language instructions into robot actions is a cornerstone of flexible robot control. We identify two key requirements to enable this: (i) universal representations that support operating diverse embodiments \citep{nair2022r3m,Ren2025MotionTA,Zheng2025UniversalAF}, and (ii) benefiting from video-language data without action labels \citep{du2023learning,Gu2023SeerLI,Black2023ZeroShotRM,Ko2023LearningTA,Cheang2025GR3TR,Lee2025MolmoActAR}. We explore their intersection, proposing \modelname, a vision–language–action framework structured as a \textit{dual-system architecture}, inspired by dual-process theories of cognition~\citep{Kahneman2011ThinkingFA} and recent hierarchical robotics frameworks~\citep{Belkhale2024RTHAH,Black20240AV,Shi2025HiRO,Nvidia2025GR00TNA,Intelligence2025pi05}. 
In our high level \textit{System 2} module, we use pixel motion as the robot action representation. We use image diffusion to learn to predict pixel motion from a single image (observation) conditioned on a language described action.
Subsequently, our embodiment-aware low level \textit{System 1} deterministically projects these action representations into executable robot actions.

We adopt pixel motion—the apparent motion of pixels between frames—as our \textit{universal motion representation}, because it is agnostic to embodiments, viewpoints, and tasks. 
By predicting pixel motion instead of full RGB images, \modelname captures essential motion patterns more efficiently (i.e. with less training data, see \Cref{subsec:rw_exp}) than text-to-video generation \citep{du2023learning,Ko2023LearningTA,Gu2023SeerLI,Black2023ZeroShotRM}. 
In contrast to operating with sparse point tracks \citep{yuan2024general,Wen2023AnypointTM,Xu2024FlowAT,Bharadhwaj2024Track2ActPP}, our dense pixel motion representation can capture both manipulator and object movements (see \Cref{fig:dense_motion}). 
Our pixel motion features also retain the inherent 2D structure of the visual domain unlike prior work modeling pixel trajectories as 1D point tracks \citep{Wen2023AnypointTM,Xu2024FlowAT}.
Moreover, dense pixel motion can be freely computed from videos using off-the-shelf algorithms like RAFT~\citep{teed2020raft}, enabling scalable, weakly supervised training on large video-caption datasets, similar to prior work on predictive world models \citep{Gu2023SeerLI,Black2023ZeroShotRM,Zhang2025DreamVLAAV}.

Optical flow, a measure of pixel motion (PM) between consecutive frames, has been leveraged to enhance motion-focused video generation \citep{Liang2024MoVideoMV, Koroglu2024OnlyFlowOF}, including in the robotics domain \citep{Gao2024FLIPFG}.
PM calculated from current and future frames is used for robot control in \citet{Ko2023LearningTA,Bharadhwaj2024Gen2ActHV}, further establishing the promise of this direction. 
In contrast, we directly generate PM from language and a single current frame (without access to future frames) using our System-2 module, offering greater data efficiency and performance (see \Cref{table:res_mw,table:real,table:real_zs}). 
Our predicted PM serves as an interpretable intermediate representation for downstream systems (e.g., our System-1), enabling even unsupervised control via hand-crafted mappings.
Alternate motion signals in image-space are used in works like \citet{Sudhakar2024ControllingTW,Shridhar2024GenerativeIA,Huang2024ReKepSR,Shi2025ZeroMimicDR}, but they rely on explicit dense annotations limiting training scalability, unlike our System-2 formulation. 
Generating PM from a single image has also been explored \citep{Walker2015DenseOF,Gao2017Im2FlowMH,Aleotti2021LearningOF}, but has been limited to the visual domain with no language conditioning. 
In contrast, our System-2 module generates PM conditioned on both visual and textual cues with no access to future frames. 

Sequences of PM generated by our System 2 are then be transformed into robot actions via \textit{System 1}, a fast and deterministic controller.
Specifically, System 1 consists of motion to action mappings that are \textit{embodiment aware}. We explore two instantiations of System 1: (a) learning mappings directly from limited expert demonstrations, and (b) hand-crafting mappings by leveraging the interpretable nature of pixel motion (motivated by \citet{Ko2023LearningTA}). 
Connecting System 1 and System 2 forms our overall language-conditioned robot control framework, \modelname. This hierarchical formulation allows operating the expensive high-level System 2 at sparse temporal intervals while invoking the lightweight low-level System 1 at dense temporal intervals for efficient inference. This also allows independent training of each system, leading to better overall training efficiency.    

In summary, our contributions are as follows:
\begin{itemize}[leftmargin=2em,noitemsep,topsep=0.0ex,itemsep=-1.0ex,partopsep=0ex,parsep=1ex]
    \item \textbf{Universal Action Representation:} 2D structured dense pixel motion as a learnable, interpretable, and manipulator-motion focused representation for robot control tasks.
    \item \textbf{Simple \& Scalable Learning:} mapping natural language actions to motion representations (pixel motion sequences) with a history-aware conditional diffusion model trained on any video-caption data, without requiring pixel-level or action trajectory annotations.
    \item \textbf{Robotics Application:} conversion of learned action representations into action policies with minimal supervision, enabling operation under zero-shot and even unsupervised settings.
\end{itemize}
We evaluate \modelname on both simulated and real-world environments, highlighting its effectiveness and generality across diverse robot control tasks.


\section{Related Work}
\label{sec:related_work}

\bhdr{Learning from Videos:} 
Robot learning has a rich history of leveraging videos to extract sub-goal information, learn strong representations, or build dynamics models for planning~\citep{lee2017learning,finn2017deep,sun2018neural,kurutach2018learning,pari2021surprising,nair2022r3m,shao2021concept2robot,chen2021learning,bahl2022human,sharma2019third,du2023learning,sivakumar2022robotic,Sudhakar2024ControllingTW,Ko2023LearningTA,Hu2024VideoPP,Ren2025MotionTA}.
Several recent works learn representations connected to language modality from video-caption data~\citep{du2023learning,Sudhakar2024ControllingTW,Ko2023LearningTA,Hu2024VideoPP}, but depend on additional action-trajectory annotations, pretrained segmentation models, or task-specific heuristics for robot control. 
We explore a similar direction, learning language-conditioned motion representations from video-caption data.  
In contrast to these works, our \modelname learns representations that are \textit{interpretable} and \textit{motion-focused}, which we use for robot control with no additional supervision. Our focus on pixel motion also allows learning more generalizable representations with less data.

\bhdr{Pixel Motion to Actions:} 
Robot navigation and control, especially in the context of aerial drones, has long benefited from optical flow representations \citep{Croon2021EnhancingOC,Lee2020AggressivePN,Hu2024SeeingTP,Argus2020FlowControlOF}, inspired by animal perception system that use optical flow for stable control and movement \citep{Gtz1968FlightCI,Arnold1974RHEOTROPISMIF,Ros2016OpticFS,Baird2021TheEO}. 
Video self-supervised learning has also extensively leveraged optical flow to learn motion representations \citep{Han2020SelfsupervisedCF,Sharma2022PixellevelCF}. 
In robot control, trajectories of pixel subsets \citep{yuan2024general,Wen2023AnypointTM,Xu2024FlowAT,Bharadhwaj2024Track2ActPP} have been used as intermediate representations, but often limit focus to specific image regions or objects, ignoring global information such as manipulator movement (e.g. see \Cref{fig:dense_motion}). 
In contrast to prior work, our \modelname models dense pixel motion (focusing on both object and manipulator movement) conditioned on textual action descriptions and visual current observations with no future frame dependency.

\begin{table}[t]
\vspace{-0.5em}
\begin{minipage}{0.48\textwidth}
\caption{
\textbf{Unique Features of \modelname.} 
\textit{Dual System}: Decoupled architecture with system 1 \& 2 modules trained separately with inference at distinct frequencies. 
\textit{Dense Motion}: Uses no heuristic / training based point sub-sampling.
\textit{2D Structure}: Pixel motion as 2D grid instead of coordinate based 1D point sequence. 
Prior work A to G are 
{\scriptsize
\citet{yuan2024general}, \citet{Gao2024FLIPFG}, \citet{Wen2023AnypointTM}, \citet{Xu2024FlowAT}, \citet{Bharadhwaj2024Track2ActPP}, \citet{Bharadhwaj2024Gen2ActHV}, \citet{Hu2024VideoPP} }
respectively. 
}
\label{tbl:related}
\vspace{-0.5em}
\centering
\small
\def\arraystretch{1.1}  
\setlength\tabcolsep{0.6em}  
\scalebox{0.84}{
\begin{tabular}{lcccccccc}
\toprule
Feature         & Ours & A & B & C & D & E & F & G \\ \midrule
Past Aware      & \cmark & \xmark & \xmark & \xmark & \xmark & \xmark & \xmark & \xmark \\
Dual System     & \cmark & \xmark & \xmark & \cmark & \cmark & \xmark & \xmark & \cmark \\
Dense Motion    & \cmark & \xmark & \cmark & \xmark & \xmark & \xmark & \xmark & \xmark \\
2D Structure    & \cmark & \cmark & \cmark & \xmark & \xmark & \xmark & \xmark & \cmark \\ 
Text Condition  & \cmark & \cmark & \cmark & \cmark & \cmark & \xmark & \cmark & \cmark \\ 
\bottomrule
\end{tabular}
}
\vspace{-0.5em}
\end{minipage}
\hspace{0.02\textwidth}
\begin{minipage}{0.48\textwidth}
    {\footnotesize 
    \hspace{1em}Prior Work \hspace{7em}  Ours \\[0.2em]
    }
    \centering
    \includegraphics[width=0.49\linewidth]{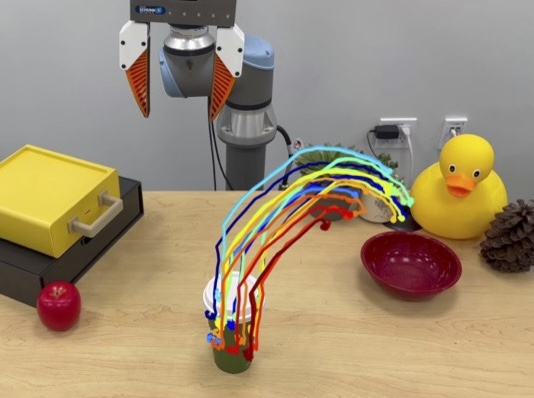}
    \includegraphics[width=0.49\linewidth]{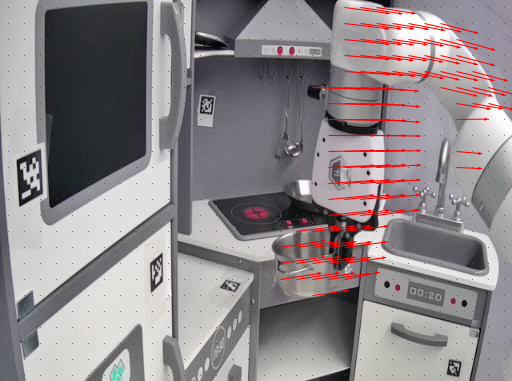}
    \vspace{-1.5em}
    \captionof{figure}{
    \textbf{Dense Motions:}
    Most prior work that use pixel trajectories focus on a subset of pixels often limited to objects of interest. The example from \citet{Xu2024FlowAT} (left) focuses on the cup movement, but ignores important action information relevant to manipulator movement. 
    In contrast, proposed \modelname generates dense pixel motions that account for both object and manipulator movements (right).
    }
    \label{fig:dense_motion}
    \vspace{-0.5em}
\end{minipage}
\end{table}

\bhdr{Diffusion-Based Motion Generation:}
Diffusion models have emerged as powerful generative frameworks capable of capturing complex data distributions through iterative denoising processes~\citep{ho2020denoising,ho2022video,ramesh2022hierarchical,zhang2023adding,singer2022makeavideo,villegas2022phenaki,ge2022long,kumari2023multiconcept,zhang2022motiondiffuse,ren2022diffusion,chen2023moddm,janner2022diffuser,du2023reduce,liu2022structdiffusion,wang2023diffusion,Chi2023DiffusionPV,Shridhar2024GenerativeIA}. 
While some works directly predict optical flow from image pairs~\citep{Saxena2023TheSE,Luo2024FlowDiffuserAO}, these tackle well-defined inputs. In contrast, \modelname generates pixel motion from a single image and language command, capturing the multi-modal nature of future motions. By also conditioning on past motion (extracted from current observations), our approach introduces temporal grounding, making it well-suited for robot control.

\bhdr{Language-Conditioned Robotic Manipulation:}
Several recent works use vision-language models for robot control ~\citep{rt1,rt2,padalkar2023open,reed2022gato,wu2023unleashing,octo_2023,driess2023palm,kim2024openvla,yuan2024robopoint,niu2024llarva,zheng2024tracevla,Li2024LLaRASR,Zawalski2024RoboticCV,Hu2024VideoPP,Sudhakar2024ControllingTW,Ko2023LearningTA,Tian2024PredictiveID,Jeong2025ObjectCentricWM} taking advantage of large-scale training with web-scale vision-language data. In contrast to prior work using sequential language models, we learn motion representations under weak supervision (only video-caption data) using zero action trajectory annotations. We also utilize an image diffusion model similar to~\citet{Hu2024VideoPP,Sudhakar2024ControllingTW,Ko2023LearningTA} but differ by learning universal and interpretable motion representations directly, which even allows conversion to robot actions directly with no further training. 



\begin{figure}[t]
    \centering
    \includegraphics[width=\linewidth]{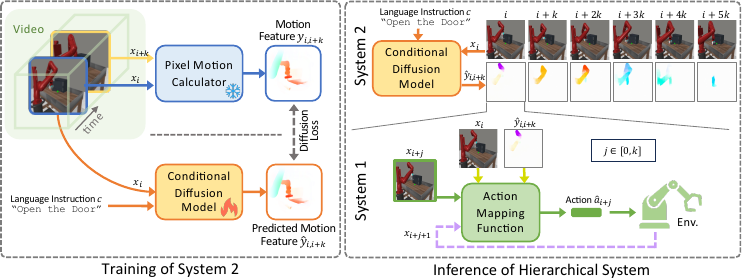}
    \vspace{-1.5em}
    \caption{
    \textbf{Overview of \modelname:}
    (Left) We learn to forecast pixel motion as universal motion features from video-caption pairs using scalable, self-supervised training of a diffusion model.
    (Right) Our \textit{System 2} forecasts motion at sparse intervals ($k$), while \textit{System 1} maps it to dense action vectors at $j$ intervals ($j < k$).    
    }
    \vspace{-0.5em}
    \label{fig:overview}
\end{figure}

\section{Methodology}
\label{sec:method}

We tackle the problem of robot control from natural language instructions by introducing a two-stage framework. Language and visual inputs are first encoded into pixel motion based representations, which are then decoded into robot actions. This dual-system architecture comprises: \textit{System 2}, a conditional image diffusion model that generates embodiment agnostic motion features at sparse temporal intervals acting as a high-level controller; and \textit{System 1}, an embodiment aware low-level controller that maps these pixel motions to executable robot action vectors. An overview of our framework, \modelname, is shown in \Cref{fig:overview}.

\subsection{System 2: Pixel Motion Forecast}
\label{subsec:system2}

Optical flow estimation from frame pairs is a well-defined problem (exact solutions exist) that has been extensively studied \citep{liu2019self,teed2020raft,xu2022gmflow,Luo2024FlowDiffuserAO}. 
In contrast, estimating pixel motion (PM) from a single image and language instruction is inherently multi-modal: a caption-frame pair may correspond to multiple valid flows, each representing a different trajectory toward the goal.
We use this challenging task as our training objective: learning a mapping from \textit{language to motion}. 
Furthermore, we incorporate temporal context by conditioning on the motion of a previous state.


Consider a video clip $\vx \in \sR^{t \times h \times w \times c}$ with $t, h, w, c$ for frames, height, width, and channels respectively. Also consider an embedding vector, $\vc$ representing the paired caption for that clip. Denoting the $i$-th frame of video as $\vx_i$, we define pixel motion, $\vy_{i, i+k}$, that corresponds to motion between frames $\vx_i \rightarrow \vx_{i+k}$ where $k$ is a constant. Our language to motion mapping function, $\gD$ becomes, 
\begin{align}
    \hat{\vy}_{i, i+k} = \gD \left( \vx_i, \vy_{i-k,i}, \vc \ | \ \theta \right)
    \label{eq:mapping}
\end{align}
where $\hat{\vy}_{i, i+k}$ is the predicted motion representation from the $i$-th state to $(i+k)$-th state \emph{without} knowing $\vx_{i+k}$ and $\theta$ are learnable parameters.

We reiterate the multi-modal output aspect of our mapping described in \Cref{eq:mapping} (i.e. one to many mapping due to multiple optimal $\hat{\vy}_{i, i+k}$). Diffusion models have shown excellent abilities to model such distributions \citep{dhariwal2021diffusion,Chi2023DiffusionPV}. Considering the 2D structure present in our images and pixel motion, for $\mathcal{D}$ we elect to utilize a 2D conditional U-Net based diffusion model \citep{ramesh2022hierarchical} operating at pixel level. 
%
Our goal is to learn a set of parameters, $\theta$ for this diffusion model based mapping as, 
\begin{align}
    \label{eq:train_obj}
    \argmin_{\theta} || \vy_{i, i+k} - 
     \gD \left( \vx_i, \vy_{i-k, i}, \vc \ | \ \theta \right) ||_2 
\end{align}
that allows our language to motion mapping to perform instruction based robot control.  
Next we dive into the learning process of our diffusion based implementation for this mapping function. 

\subsection{Diffusion based Motion Representation Learning}
\label{subsec:ssl}

\bhdr{Background:} 
Diffusion Models generate data by progressively denoising corrupted signals, optionally conditioned on a goal input. While inference follows this iterative refinement process, training is conducted more efficiently using parallel denoising steps: the model is trained to predict less noisy versions of intermediate corrupted signals generated from clean data, a procedure analogous to teacher forcing (more details in \Cref{app:dm_detail}).

\begin{figure}[t]
    \centering
    \includegraphics[width=\linewidth]{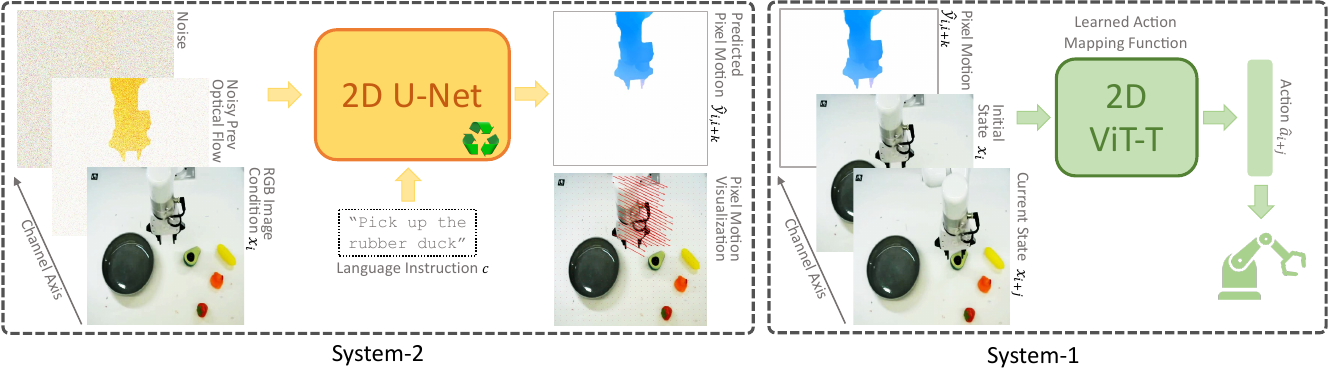}
    \vspace{-1.5em}
    \caption{
    \textbf{\modelname Architecture:}
    (Left) Diffusion model generates pixel motion conditioned on RGB image, prior motion, and caption. Visualized predictions are overlaid as arrows. 
    (Right) ViT-T network maps predicted motion to robot actions in supervised setting, conditioned on initial/current states and target motion.
    }
    \label{fig:method}
    \vspace{-1.0em}
\end{figure}

\bhdr{Architecture:}
The defacto architecture for diffusion based conditional image generation is the 2D conditional U-Net \citep{Ronneberger2015UNetCN}, which maps between 2D RGB images with an embedding based conditioning through cross-attention in the model intermediate layers. 
Basing off this setup, we modify the input and output heads to process 7 and 2 channel tensors respectively (instead of default 3 channel RGB). Two of the input channels and the two output channels correspond to our pixel motion target (noise input and clean output). 
The remaining 5 input channels correspond to our 2D-structured conditions:
previous pixel motion (2 channels) and current state image (3 channels).
These conditional inputs are not subject to the standard noise corruption schedule during training or inference (details in \Cref{app:dm_detail}). 
The textual embedding is provided as the default embedding condition. 
Our channel modification to accommodate additional structured conditions allows a minimal design, retaining the general structure of the U-Net that is known to excel at 2D generative modeling.   
Such input channel concatenation based conditioning has been used in diffusion literature for different tasks \citep{Saxena2023TheSE,ho2022video} and is inspiration for our design.
We illustrate this architecture in \Cref{fig:method} (left). 

\bhdr{Calculating Pixel Motion Ground-truth:}
We utilize the RAFT algorithm \citep{teed2020raft} to calculate our target pixel motion $\vy_{i, i+k}$, using frames $\vx_i$ and $\vx_{i+k}$. 
This is an efficient iterative algorithm that calculates a good estimate of optical flow, in other words, pixel motion. Each pixel motion, $\vy_{i, i+k} \in \sR^{h \times w \times 2}$, contains two channels for spatial directions, that are normalized to a $(0, 1)$ range. All motion is represented within this 2D space - extensions to a third depth dimension are left as a future direction. Our experiments indicate the sufficiency of such 2D spaces to encode motions relevant to robot actions. 
We note that given the presence of background motions in both natural and simulation images (e.g. shadows moving with objects), this target pixel motion contains noise that is not directly relevant to the underlying motion, underscoring the challenging nature of our self-supervision objective.    

\bhdr{Previous Pixel Motion Representation:}
The other input signal to our mapping function is past pixel motion. Motivated by success of teacher forcing both language \citep{Radford2019LanguageMA} and video \citep{Song2025HistoryGuidedVD} generation, we use the target pixel motion of previous time steps during our System-2 training. 
We also note the importance of representing pixel motion relative to current state as our mapping function is conditioned on the current image (details in \Cref{app:relative_of}). Similar findings are observed in image-pair based optical flow calculation literature \citep{Ko2023LearningTA}.  

\bhdr{Language Instruction Embeddding:}
The primary input conditioning of our mapping function is the natural language based action description that is used to control the generated motions. Following prior robotics literature \citep{padalkar2023open}, we use a Universal Sentence Encoder model \citep{Cer2018UniversalSE} to convert textual instructions to fixed size embedding vectors. This embedding model is trained to capture sentence level meanings. We use an off-the-shelf pretrained version, keeping all model parameters unchanged (more details in \Cref{app:lang_embed}).

\bhdr{Training:}
Our training uses the standard diffusion denoising objective \citep{ho2020denoising} between predicted ($\hat{\vy}_{i, i+k}$) and target ($\vy_{i, i+k}$) pixel motion. The conditional 2D inputs, $\vx_i$ and $\vy_{i-k, i}$ are not subject to a noising schedule. The image condition, $\vx_i$, remains uncorrupted while the previous pixel motion, $\vy_{i-1,i}$, is set to random noise or a partially corrupted version to align with inference settings. We also introduce zero motion to ends of videos such that when textual instruction is complete, those visual states map to zero motion. More details in \Cref{app:dm_detail}. 

\bhdr{Inference:}
We forecast pixel motion from $i$ to $i+k$ timestamp using a 25-step DDIM schedule with only the current image observation $\vx_i$. At the initial step, the model only takes the image $\vx_i$ (state observation), language instruction $c$, and zero vector as the previous pixel motion. For subsequent steps, previous motion is calculated from the past-current observation pair, enabling sequential pixel motion generation that drives the system toward fulfilling the language command.

\subsection{System 1: Pixel Motion to Action Mapping}
\label{subsec:system1}

Our System 2 produces pixel motion conditioned on a given state-instruction pair.
We next detail how these pixel motion representations are mapped into action vectors that directly control the robot. 
Consider a mapping function, $\gF$, operating at dense temporal intervals:
\begin{align}
    \label{eq:actions}
    \hat{\va}_{i+j} = \gF \left( \hat{\vy}_{i, i+k}, \vx_i, \vx_{i+j} \right),
\end{align}
where $ j \in \left[0,k\right]$, $i$ is a multiple of $k$ (for a hyperparameter $k$), and $\hat{\va}_{i+j}$ denotes the predicted action vector for the $(i+j)$-th state. 
An overview of this formulation is shown in \Cref{fig:overview} (right).

While \textit{System 2} is trained as a general-purpose motion generator across diverse embodiments, viewpoints, and environments, 
action vectors $\va_i$ are inherently embodiment-specific.
Hence, we design \textit{embodiment-aware} mapping functions to serve as \textit{System 1 (Action Mapping)}, that are capable of converting pixel motion into executable robot actions.

\bhdr{Learned Mapping:}
We implement a neural network-based mapping function that can be trained using ground-truth action trajectories. 
Given the 2D spatial structure of the inputs to $\gF$ (i.e., $\hat{\vy}_{i,i+j}$, $\vx_i$, $\vx_{i+j}$), we channel-concatenate them and feed the resulting tensor to a lightweight vision transformer to predict action vectors.
This architecture is illustrated in \Cref{fig:method} (right).
The network is trained on a limited amount of embodiment-specific demonstration data. 
Connecting this learned \textit{System 1} with \textit{System 2} following \Cref{eq:actions}, we obtain a complete pipeline for language-conditioned robot control.
We refer to the resulting system, which uses a supervised learned mapping, as \modelshort-S.

\bhdr{Hand-Crafted Mapping:}
The interpretable nature of pixel motion also enables hand-crafted designs for $\gF$.
We refer to the resulting pipeline based on hand-crafted mappings as \modelshort-H.
For simulated environments where ground-truth segmentations and depth maps are available, we follow the methodology in~\citep{Ko2023LearningTA} to define action mappings, ensuring a fair evaluation of the utility of our pixel motion predictions compared to prior works.
For real-world robot control, we construct viewpoint-specific hand-crafted mappings following~\citep{Li2024LLaRASR}.
Further details on both learned and hand-crafted mappings are provided in \Cref{app:handcraft_map}.

We highlight how our System 1 operates at a frequency different to our System 2, allowing a balance between efficiency and dense control. Our System 1 is also designed to be lightweight, given how it performs an almost deterministic mapping.  



\section{Experimental Results}
\label{sec:result}

We conduct experiments on 15 task styles spanning both simulated and real-world environments to highlight the strong performance of our proposed \modelname framework. 
We also present multiple ablations to justify key design choices within our method.

\bhdr{Implementation Details:}
Our framework consists of \textit{System 2 (Motion Generation)} containing a diffusion model, and \textit{System 1 (Action Mapping)} containing either a learned or hand-crafted mapping function.
We pretrain the diffusion model on a subset of the OpenX dataset~\citep{padalkar2023open}, followed by optional fine-tuning on downstream task datasets. 
Pretraining is performed for 300,000 iterations with a learning rate of $1\text{e-}4$, following a cosine learning rate schedule with 500 warmup steps, using 8 A100 GPUs (48GB) with a per-device batch size of 32 samples. 
Fine-tuning is performed for 100,000 iterations on 4 A5000 GPUs (24GB) with a batch size of 32 and a learning rate of $1\text{e-}5$, again following a cosine schedule with 500 warmup steps.
The learned action mapping (System 1) is trained separately using a vision transformer for 10,000 iterations on a single A5000 GPU with a batch size of 128 and a learning rate of $1\text{e-}4$.
During inference of our System 2 diffusion model, we use a DDIM scheduler with 25 steps to generate flow sequences, starting from noise. For each invocation of System 2, we run System 1 for 10 control steps (or until convergence in the hand-crafted setting). This hierarchical procedure is repeated until the episode terminates.

\begin{table}[t]
\centering
\small 
\begin{minipage}{0.48\textwidth}
    \caption{
    \textbf{Zero-Shot Transfer on Real World Tasks:} 
    We directly deploy our pretrained model (with no fine-tuning) on real world tasks. 
    We highlight our strong performance compared to baselines in this highly challenging setting. 
    Evaluations follow \cite{Li2024LLaRASR}.
    }
    \label{table:real_zs}
    \vspace{-1.0em}
    \def\arraystretch{1.3}  
    \setlength\tabcolsep{0.5em}  
    \scalebox{0.85}{
    \begin{tabular}{lcccccc}
    \toprule
    Method  & \makecell{Video Only\\ Training} & T1 & T2 & T3 & T4 & Avg \\ \midrule
    RT-2 Style   & \xmark &   0 &  0 & 0  & 0  &  0   \\
    LLaRA        & \xmark &  40 & 20 & 10 & 20 & 22.5 \\
    AVDC         & \cmark &  0 &  0 & 0  & 0  &  0 \\
    GPT-4o       & \cmark &  20 & 30 & 10 & 15 & 18.8 \\ \rowcolor{Gray} 
    LTM-H (ours) & \cmark &  40 & 30 & 35 & 30 & 33.8 \\ \bottomrule
    \end{tabular}}
\end{minipage}
\hspace{0.02\textwidth}
\begin{minipage}{0.48\textwidth}
    \caption{
    \textbf{Finetuned on Real World:} 
    \modelname benefits from both robot (RD) and human (HD) demonstrations highlighting the embodiment agnostic nature of our Sys-2, in contrast to prior work.
    }
    \label{table:real}
    \vspace{-1.0em}
    \def\arraystretch{1.3}  
    \setlength\tabcolsep{0.6em}  
    \scalebox{0.80}{
    \begin{tabular}{lcccccc}
    \toprule
    Method       & Data  & T1 & T2 & T3 & T4 & Average \\ \midrule
    RT-2 Style   & -     &   0 &  0 & 0  & 0  &  0   \\
    LLaRA        & -     &  70 & 80 & 55 & 55 & 65.0 \\ 
    AVDC         & RD    & 10 & 20 & 0  &  0 &  7.5 \\ 
    AVDC         & RD+HD &  0 &  0 & 0  &  0 &  0.0 \\ 
    LTM-H (ours) & HD    & 40 & 35 & 40 & 30 & 36.3 \\ 
    LTM-H (ours) & RD    & 80 & 70 & 65 & 60 & 68.8 \\ \rowcolor{Gray} 
    LTM-H (ours) & RD+HD & 80 & 75 & 65 & 65 & 71.3 \\ 
    \bottomrule
    \end{tabular}}    
\end{minipage}
\vspace{-0.5em}
\end{table}

\begin{figure*}[t]
    \centering
    \begin{minipage}{0.03\linewidth}
    \centering
    \begin{tabular}{c}
         \rotatebox{90}{\texttt{Human}}   \\[1.8em] 
         \rotatebox{90}{\texttt{Robot}}   
    \end{tabular}
    \end{minipage}
    \begin{minipage}{0.96\linewidth}
    \begin{minipage}{\linewidth}
        \centering\centering
        \includegraphics[width=\linewidth]{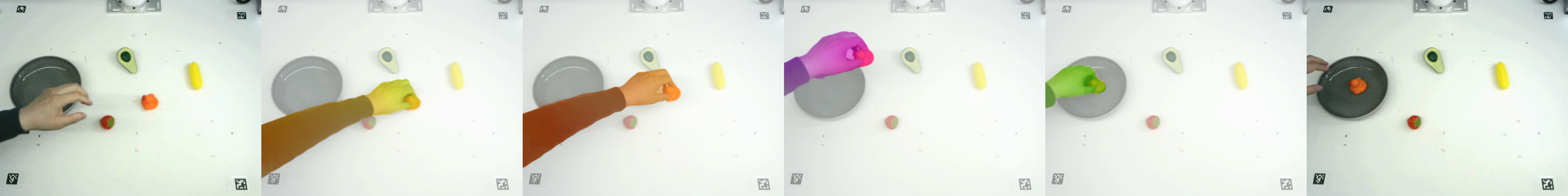}
    \end{minipage}
    \begin{minipage}{\linewidth}
        \centering\centering
        \includegraphics[width=\linewidth]{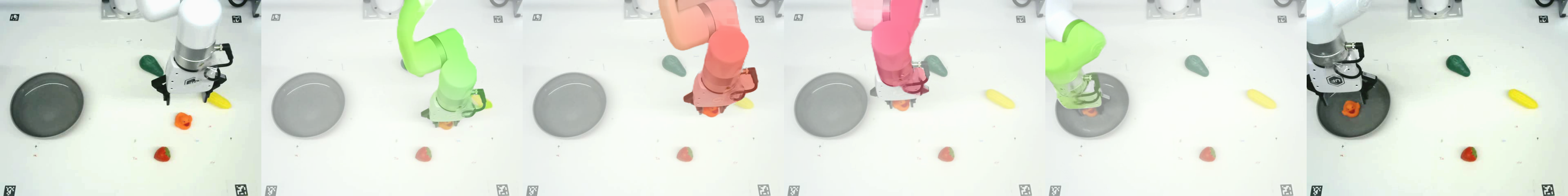}
    \end{minipage}
    \end{minipage}
    \vspace{-0.5em}
    \caption{
    \textbf{Human (HD) \& Robot (RD) Demonstrations:} 
    We visualize frames from two sample demonstrations on our real world environment. Pixel motion overlaid on intermediate frames. These human (top) and robot (bottom) demonstrations can both be used to fine-tune our System-2, highlighting a unique aspect of \modelname. Both examples use the same caption \texttt{"Pick up the rubber duck and place on the bowl."}
    }
    \label{fig:demos}
    \vspace{-0.5em}
\end{figure*}

\subsection{Real-World Environment}
\label{subsec:rw_exp}

We first evaluate on four styles of real world tasks using the xArm Table Top environment constructed following ~\citet{Li2024LLaRASR}. We select this environment and task styles for its ease of fair comparison to prior work, interpretable action dynamics (each state suggests a clear next motion), and demands for visual grounding, semantic understanding, and distractor robustness. 
The tasks involve object manipulations specified by language commands (details in \Cref{app:real_world}).

\bhdr{Training \& Evaluation:}
We train \textit{System 2} on the OpenX subset, followed by optional fine-tuning on demonstrations from the real-world environment. We collect 10 robot demonstrations (RD) and 50 human demonstrations (HD) per task style.
We replicate AVDC \citep{Ko2023LearningTA} by training under identical conditions. 
All other baselines are implemented following settings from \citet{Li2024LLaRASR}.
For \textit{System 1}, we construct a hand-crafted mapping function combining ideas from \citet{Ko2023LearningTA,Li2024LLaRASR} (details in \Cref{app:real_world}).
We follow evaluation settings identical to \citet{Li2024LLaRASR}, evaluating each policy across 4 task styles with a fixed camera view and 20 randomized trials per task style.
Each trial uses different initial positions of the objects present in the environment.

\bhdr{Zero-Shot Results:}
We present results for zero-shot evaluation in \Cref{table:real_zs}. Strong performance of this \textit{pre-trained only} Sys-2 module highlight the significance of our large-scale pretraining.

\bhdr{Finetuning Results:}
We next fine-tune our Sys-2 module on the robot (RD) and human (HD) demonstrations, presenting these results in \Cref{table:real}. 
Compared to AVDC \citep{Ko2023LearningTA} that makes RGB predictions, our Sys-2 module that predicts pixel motion benefits from human demonstrations ($+2.5$\%). 
In contrast, AVDC predictions break down when trained on both RD and HD. We attribute this to the greater difference between human vs robot manipulators in RGB space compared to pixel motion space (see motion overlay in \Cref{fig:demos}, distribution analysis in \Cref{app:dist_analysis}, and \cite{Xu2024FlowAT}).
We also highlight that fine-tuning here uses only video-caption pairs and no action ground-truth. This is what allows learning from both RD and HD data. 
Collection of such human demonstrations (HD) is much faster compared to teleoperated robot demonstrations (RD) \citep{Cheang2025GR3TR}, underscoring the value of our \modelname framework.

\subsection{MetaWorld Simulated Environment}

Our second set of evaluations use 11 tasks from the MetaWorld~\citep{yu2019meta} simulated environment containing a Sawyer robot arm, constructed following \citep{Ko2023LearningTA}. We select this environment and tasks for direct comparison to \citep{Ko2023LearningTA}, which is the closest prior work to our method, that similarly uses dense pixel motion for robot manipulation. 
These tasks also span key challenges in robot control such as complex 3D motions (e.g. button-press-top), contact-rich manipulation (e.g. basket-ball), and semantic understanding (e.g. door open vs close). 
Each task episode corresponds to successfully completing an action described in natural language.

\bhdr{Training:}
We pretrain \textit{System 2} on the OpenX subset, followed by additional training on 165 MetaWorld videos (identical to the split used in \citet{Ko2023LearningTA}).
For the learned variant of \textit{System 1}, we train on 20 expert demonstrations per task.
We also implement a hand-crafted variant of System 1, following the design in \citet{Ko2023LearningTA} to ensure fair comparison.

\bhdr{Baselines:}
All baselines follow settings in \citet{Ko2023LearningTA}). The behaviour cloning (BC) baselines are trained on 15,216 labeled frame-action pairs (over 5x more data). BC-Scratch uses a randomly initialized ResNet-18 while BC-R3M uses pretrained weights from \citet{nair2022r3m}. Diffusion Policy follows settings in \citet{Chi2023DiffusionPV} and is trained on the same data. UniPi \citep{du2023learning} uses the outputs of the AVDC model and its predictor is trained on the same data used for BC baselines. 
AVDC (Flow \& Default) are trained identical to \citet{Ko2023LearningTA} using same 165 Metaworld videos. AVDC (PT) is additionally pretrained on our OpenX subset making the training identical to \modelshort. 
Im2Flow2Act \citep{Xu2024FlowAT} and ATM \citep{Wen2023AnypointTM} follow their default implementations and are also trained identically using the same training data as \modelname. 

\bhdr{Evaluation:}
Following evaluation settings identical to~\citep{Ko2023LearningTA}, we evaluate each policy across 11 tasks.
For each task, videos are rendered from 3 distinct camera poses, with 25 randomized trials (different initial positions of the robot arm and objects) for each view.

\bhdr{Results:}
We present the success rates for the 11 tasks and the average across tasks in \Cref{table:res_mw}.
Notably, several prior works~\citep{du2023learning,Ko2023LearningTA} exhibit moderate success rates, underscoring the difficulty of the benchmark.
Both our \modelshort-H and \modelshort-S variants achieve strong overall performance, highlighting the effectiveness of our framework.
Our approach of directly predicting pixel motion compared to RGB in AVDC \citep{Ko2023LearningTA} achieves clear performance improvements ($+9.0$\%). Moreover, AVDC fails to benefit from pretraining, which we attribute to the greater domain gap across embodiments in RGB space compared to pixel motion space.  
Another important point of comparison is the AVDC (flow) baseline, which also uses pixel motion prediction but differs in model architecture, flow representation, and training procedures.
We attribute this improved performance of \modelname over AVDC to our unique design choices.
In comparison to ATM \citep{Wen2023AnypointTM}, our improved performance highlights the usefulness of our dense pixel motion features. 

\begin{table}[t]
\caption{
\textbf{Results on MetaWorld Environment:}
We report the mean success rate across tasks. Each entry of the table shows the average success rate aggregated from $3$ camera poses with $25$ seeds for each camera pose.}
\label{table:res_mw}
\vspace{-0.5em}
\centering
\small
\def\arraystretch{1.2}  
\setlength\tabcolsep{0.5em}  
\scalebox{0.90}{
\begin{tabular}{lcccccccccccc}
\toprule
& {\footnotesize \rotatebox{70}{door-open}} 
& {\footnotesize \rotatebox{70}{door-close}} 
& {\footnotesize \rotatebox{70}{basketball}} 
& {\footnotesize \rotatebox{70}{shelf-place}} 
& {\footnotesize \rotatebox{70}{btn-press}} 
& {\footnotesize \rotatebox{70}{btn-top}} 
& {\footnotesize \rotatebox{70}{faucet-close}} 
& {\footnotesize \rotatebox{70}{faucet-open}} 
& {\footnotesize \rotatebox{70}{handle-press}} 
& {\footnotesize \rotatebox{70}{hammer}} 
& {\footnotesize \rotatebox{70}{assembly}} 
& {\footnotesize \rotatebox{70}{Overall}} \\
\midrule
BC-Scratch     & 21.3 & 36.0 & 0.0 & 0.0 & 34.7 & 12.0 & 18.7 & 17.3 & 37.3 & 0.0 & 1.3 & 16.2 \\
BC-R3M         & 1.3 & 58.7 & 0.0 & 0.0 & 36.0 & 4.0 & 18.7 & 22.7 & 28.0 & 0.0 & 0.0 & 15.4 \\
Diffusion Policy & 45.3 & 45.3 & 8.0 & 0.0 & 40.0 & 18.7 & 22.7 & 58.7 & 21.3 & 4.0 & 1.3 & 24.1 \\
UniPi (With Replan) & 0.0 & 36.0 & 0.0 & 0.0 & 6.7 & 0.0  & 4.0 & 9.3 & 13.3 & 4.0 & 0.0 & 6.1 \\
Im2Flow2Act    & 0.0 & 0.0 & 0.0 & 4.0 & 6.3 &  0.0 &  7.3 &  4.7 & 0.0 & 0.0 & 0.0 & 2.0 \\
ATM            & 75.3 & 90.7 & 24.0 & 16.3 & 77.3 & 76.7 & 50.0 & 62.7 & 92.3 & 4.3 & 2.0 & 52.0 \\
AVDC (Flow)    & 0.0 & 0.0 & 0.0 & 0.0 & 1.3 & 40.0 & 42.7 & 0.0 & 66.7 & 0.0 & 0.0 & 13.7 \\
AVDC (Default) & 72.0 & 89.3 & 37.3 & 18.7 & 60.0 & 24.0 & 53.3 & 24.0 & 81.3 & 8.0 & 6.7 & 43.1 \\ 
AVDC (PT)      & 72.0 & 88.7 & 37.3 & 18.7 & 58.7 & 24.3 & 53.3 & 24.0 & 81.3 & 8.0 & 6.7 & 42.9 \\ \rowcolor{Gray}
LTM-H (Ours)    & 76.0 & 94.7 & 38.0 & 15.3 & 82.0 & \textbf{84.7} & 41.3 & 33.3 & 97.3 & 4.3 & 6.7 & 52.1 \\ \rowcolor{Gray}
LTM-S (Ours)    & \textbf{77.3} & \textbf{95.0} & \textbf{39.0} & \textbf{20.3} & \textbf{82.7} & 84.3 & \textbf{52.3} & \textbf{68.3} & \textbf{98.0} & \textbf{10.3} & \textbf{7.7} & \textbf{57.7} \\
\bottomrule
\end{tabular}
}
\end{table}

\subsection{Ablation Studies}
\label{subsec:ablation}

We conduct a series of ablative studies with LTM-S on the MetaWorld benchmark to evaluate the importance of key components within \modelname. Results are summarized in \Cref{table:ablate}.

\bhdr{System 2 Input Conditioning \& Pretraining:}
Removing visual (``Img"),  language (``Lang"), or history information (``Prev Flow") from conditional inputs to the diffusion model significantly reduces performance, highlighting importance of each conditioning signal.
On the other hand, removing diffusion model pretraining (``PT'') leads to a modest performance drop, indicating that while pretraining aids convergence and performance, the framework remains effective with limited finetuning alone.

\bhdr{Simpler Baselines:}
Replacing diffusion (``No diffusion") with an autoencoder breaks System-2 learning process. We believe diffusion is more suited for learning the multi-modal output-space of language to motion distributions. 
Modifying conditioning strategy to cross-attention (``CA instead of concat") also degrades performance. 
We attribute this to loss of spatial information when performing cross-attention with spatially-averaged visual embeddings. 
Skipping the iterative System-1 design (running System-1 at same frequency), and generating multiple actions per System-2 generated motion at once (``Sys-1 \& 2 same freq") also degrades success rates, validating our design choices.
Additionally, bypassing intermediate motion representations (``Only learned Sys-1") leads to poor results, underscoring the clear role played by our System-2 module. See \Cref{app:ablate} for a detailed discussion.

\begin{table}[t]
\centering
\small
\begin{minipage}{0.38\textwidth}
\caption{
    \textbf{Ablation Study:} 
    We report mean success rate \%  (overall) on MetaWorld benchmark with our \modelshort-S variant. (left) Results highlight importance of key components in our System-2 model. 
    (right) Results justify several high-level design choices of our framework.
}
\label{table:ablate}
\vspace{-1.0em}
\end{minipage}
\hspace{0.01\textwidth}
\begin{minipage}{0.30\textwidth}
    \def\arraystretch{1.3}  
    \setlength\tabcolsep{0.6em}  
    \scalebox{0.75}{
    \begin{tabular}{ccccc}
    \toprule
    Img  & Lang & Prev Flow & PT & Overall \\ \midrule \rowcolor{Gray}
    \cmark & \cmark & \cmark & \cmark & 57.7 \\
    \cmark & \cmark & \cmark & \xmark & 53.1 \\
    \cmark & \cmark & \xmark & \xmark & 50.2 \\
    \cmark & \xmark & \xmark & \xmark & 39.7 \\
    \xmark & \cmark & \xmark & \xmark &  5.6 \\
    \bottomrule
    \end{tabular}}    
\end{minipage}
\hspace{0.02\textwidth}
\begin{minipage}{0.26\textwidth}
    \def\arraystretch{1.3}  
    \setlength\tabcolsep{0.6em}  
    \scalebox{0.75}{
    \begin{tabular}{lc}
    \toprule
    Method  & Overall \\ \midrule \rowcolor{Gray}
    Ours (default)              & 57.7 \\
    No diffusion                & 16.2 \\
    CA instead of concat        & 15.8 \\
    Sys-1 \& 2 same freq       & 48.7 \\
    Only learned Sys-1          & 3.2 \\
    \bottomrule
    \end{tabular}}     
\end{minipage}
\end{table}


\section{Conclusion}
\label{sec:conclusion}

We presented \modelname, a scalable vision-language-action framework that decouples motion generation and action execution through a dual-system architecture. 
By leveraging diffusion models to learn universal pixel motion representations from video-caption data, our \textit{System 2} enables generalizable, interpretable motion planning without dense supervision.
These motions are translated into robot actions by our embodiment-aware \textit{System 1}, using either learned or hand-crafted mappings.
Extensive experiments across simulated and real-world environments demonstrate strong performance of \modelname, highlighting the promise of universal motion representations as a bridge between language, vision, and action for scalable robot learning.

\subsection*{Limitations}

LangToMo is pretrained on large-scale video-caption data, but relies on hand-crafted or learned action mappings in System 1 which can be costly for each new embodiment. Learning robust, transferable mappings remains an open challenge.
Also, our framework models motion using 2D pixel motions, which currently lacks depth cues. Extending to 3D motion representations is left as a future direction.
In terms of speed, despite operating at sparse intervals, System 2 relies on diffusion models that remain computationally expensive at inference time, limiting use in resource-constrained deployments. This is another future direction we hope to explore further. 
Finally, we currently do not account for ego motion in training videos: we limit our training to fixed camera videos (no ego motion). A key next direction is extending our System-2 training to include videos with ego motion, which would allow scaling to any kind of video.

\clearpage

\bibliographystyle{iclr2025_conference}
\bibliography{main}

\clearpage

\vspace{5em}
{\Large \textbf{Appendix}}
\appendix

\section{Dataset Details}
We use a subset of OpenX for pretraining of our System-2 module. We use \texttt{fractal}, \texttt{taco\_play}, \texttt{language\_table}, \texttt{stanford\_hydra}, \texttt{ucsd\_pick\_place}, \texttt{cmu\_pickup}, and \texttt{utaustin\_mutex} datasets from the OpenX collection. Frame sampling is performed uniformly to maintain a fixed, common action count between frames across each dataset by normalizing for control frequency. 
We present details of each subset in \Cref{app:tbl_data}.

\begin{table}[ht]
\centering
\small
\caption{
\textbf{Pretraining Dataset:}
We use 7 sub-datasets from the OpenX collection for pretraining of our System-2 module. Note that training is performed jointly with 3 different embodiments operated at different control frequencies. Our pixel motion based representations allows training jointly with such data using a common training objective across data from all embodiments. 
}
\label{app:tbl_data}
\def\arraystretch{1.3}  
\setlength\tabcolsep{0.8em}  
\scalebox{0.90}{
\begin{tabular}{l|c|c|c|c}
\toprule
Dataset         & Control Frequency & Episodes & Size (GB) & Robot \\ \midrule
\texttt{fractal}           & 3 & 73,499 & 111.06 & Google Robot \\
\texttt{taco\_play}        & 15 & 3,242 & 47.77 & Franka \\
\texttt{language\_table}   & 10 & 442,226 & 399.22 & xArm \\
\texttt{stanford\_hydra}   & 10 & 550 & 72.48 & Franka \\
\texttt{ucsd\_pick\_place} & 3 & 1,355 & 3.53 & xArm \\
\texttt{cmu\_pickup}       & 20 & 520 & 50.29 & Franka \\
\texttt{utaustin\_mutex}   & 20 & 1,500 & 20.79 & Franka \\ \bottomrule
\end{tabular}}
\end{table}

\section{Additional Experimental Results}
\label{app:extra_exp}

We present more experiments on our real world environment as well as two additional simulation environments to further investigate behaviour of our \modelname framework.

\subsection{Semantic Awareness}

\begin{figure}[ht]
\begin{minipage}{\linewidth}
    \centering
    \includegraphics[width=0.24\linewidth,height=6.5em]{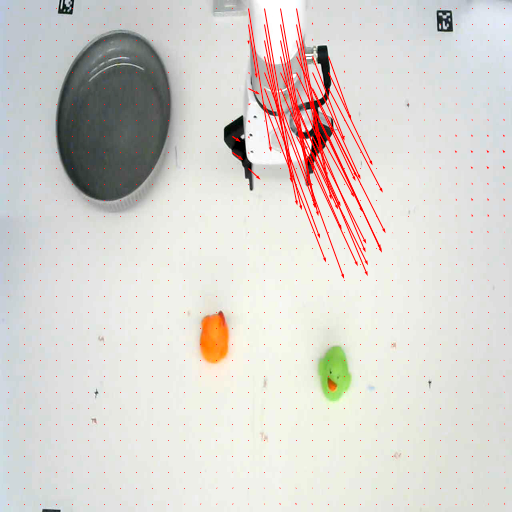}
    \includegraphics[width=0.24\linewidth,height=6.5em]{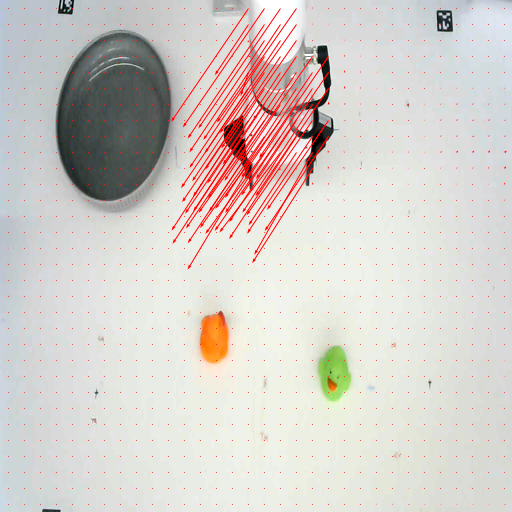}
    \includegraphics[width=0.24\linewidth,height=6.5em]{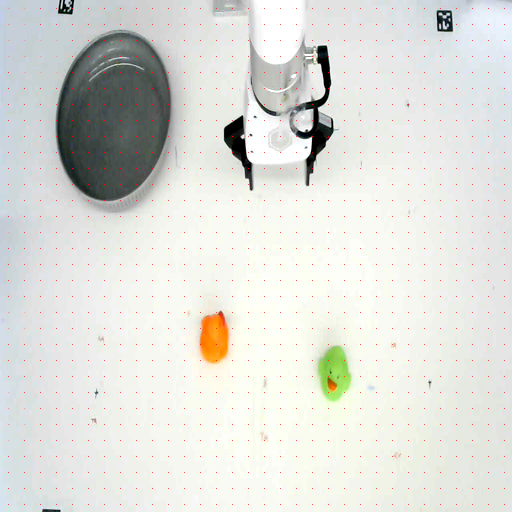}
    \includegraphics[width=0.24\linewidth,height=6.5em]{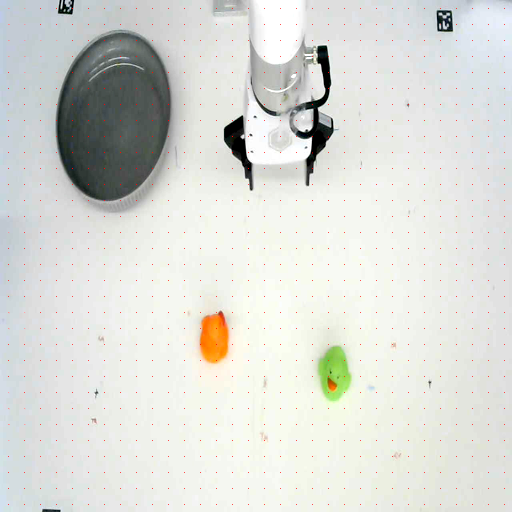}
\end{minipage}
\begin{minipage}{\linewidth}
\vspace{-1.8em}
{\tiny
\hspace{0.4em}
\textls[-10]\texttt{Pick up the green duck}  \hspace{6.3em} 
\textls[-10]\texttt{Pick up the orange duck} \hspace{6.1em} 
\textls[-10]\texttt{Pick up the green duck}  \hspace{6.2em} 
\textls[-10]\texttt{Pick up the orange duck}
}
\end{minipage}
\begin{minipage}{\linewidth}
\vspace{-2.6em}
{\tiny
\hspace{0.4em}
\textls[-10]\texttt{(Ours)} \hspace{12.9em}
\textls[-10]\texttt{(Ours)} \hspace{13.0em}
\textls[-10]\texttt{(AVDC)} \hspace{12.1em} 
\textls[-10]\texttt{(AVDC)}
}
\end{minipage}
\vspace{-3.0em}
\caption{
\textbf{Semantic Awareness Visualization:}
We visualize outputs from our System-2 module (ours; left two figures) for two examples containing the same starting state (visual observation) but different action goals (textual command).  
\modelname generates meaningful motions for scenarios needing semantic understanding. 
We also compare against the AVDC baseline \citep{Ko2023LearningTA} (trained on data identical to our \modelname)
that generates next frame RGB images instead of pixel motion. 
For both cases, AVDC generates the same input frame as its output, i.e. a static next image, seemingly disregarding the language command.
}
\label{fig:semantics}
\end{figure}

In \modelname, our System-2 module plays the role of understanding semantics within the action goal (textual command) and converting it into meaningful pixel motion representations. 
We empirically validate this functionality by visualizing two examples that contain the same visual observation but different action goals. We illustrate this in \Cref{fig:semantics}.
Our \modelname System-2 module shows much better language awareness in comparison to the AVDC baseline \citep{Ko2023LearningTA}. 

\subsection{CALVIN Evaluation}

CALVIN \citep{Mees2021CALVINAB} is another simulation benchmark used in several recent works such as \citet{Hu2024VideoPP}. We evaluate our model on this benchmark following settings in \citet{Hu2024VideoPP} and summarize these results in \Cref{tbl:calvin}. 
All prior work numbers are directly borrowed from \citet{Hu2024VideoPP} since we follow their exact settings for evaluation. We explore the two settings of training on the full ABC split and 10\% of the ABC split. Evaluation is always performed on the unseen D split. 
Each task is a set of five sequential sub-tasks and we use the task success rates along with average length metrics for evaluation similar to \citet{Hu2024VideoPP}. 

In the first case (100\% ABC), we perform competitively outperforming several recent works. 
Concurrent works, VPP \citep{Hu2024VideoPP} and DreamVLA \citep{Zhang2025DreamVLAAV} outperform us on this split. We note that both these models are pretrained on significantly more data than our \modelshort model. VPP also uses a larger sized model (1.5B) compared to our \modelshort (0.86B). 
In the second case, (10\% ABC), we outperform VPP and GR-1 \citep{wu2023unleashing} highlighting the data efficiency aspect of our \modelname framework. 

\begin{table}[t]
\caption{
\textbf{CALVIN Evaluation:}
Zero-shot long-horizon evaluation on the Calvin ABC$\rightarrow$D benchmark where agent is asked to complete five chained tasks sequentially based on instructions. 
}
\centering
\small
\def\arraystretch{1.1}  
\setlength\tabcolsep{0.80em}  
\scalebox{0.9}{
\begin{tabular}{cccccccc}
\toprule
\multirow{2}{*}{\textbf{Method}}& 
\multirow{2}{*}{\textbf{Training Data}} 
& \multicolumn{5}{c}{\textbf{$i^{th}$ Task Success Rate $\uparrow$}} 
& \multirow{2}{*}{\textbf{Avg. Len $\uparrow$}}
\\ \cline{3-7} 
& & \textbf{1} & \textbf{2} & \textbf{3} & \textbf{4} & \textbf{5} & \\ \midrule
RT-1             & 100\% ABC & 0.533 & 0.222 & 0.094 & 0.038 & 0.013 & 0.90 \\
Diffusion Policy & 100\% ABC & 0.402 & 0.123 & 0.026 & 0.008 & 0.00 & 0.56 \\ 
Robo-Flamingo    & 100\% ABC & 0.824 & 0.619 & 0.466 & 0.331 & 0.235 & 2.47 \\
Uni-Pi           & 100\% ABC & 0.560 & 0.160 & 0.080 & 0.080 & 0.040 & 0.92 \\
MDT              & 100\% ABC & 0.631 & 0.429 & 0.247 & 0.151 & 0.091 & 1.55 \\
Susie            & 100\% ABC & 0.870 & 0.690 & 0.490 & 0.380 & 0.260 & 2.69 \\
GR-1             & 100\% ABC & 0.854 & 0.712 & 0.596 & 0.497 & 0.401 & 3.06 \\ 
Vidman           & 100\% ABC & 0.915 & 0.764 & 0.682 & 0.592 & 0.467 & 3.42 \\ 
RoboUniview      & 100\% ABC & 0.942 & 0.842 & 0.734 & 0.622 & 0.507 & 3.65 \\ 
VPP              & 100\% ABC & 0.965 & 0.909 & 0.866 & 0.820 & 0.769 & 4.33 \\ 
DreamVLA         & 100\% ABC & 0.982 & 0.946 & 0.895 & 0.834 & 0.781 & 4.44 \\ \rowcolor{Gray}
LTM-S (ours)     & 100\% ABC & 0.971 & 0.824 & 0.728 & 0.672 & 0.606 & 3.81 \\ \midrule
GR-1             &  10\% ABC & 0.672 & 0.371 & 0.198 & 0.108 & 0.069 & 1.41 \\
VPP              &  10\% ABC & 0.878 & 0.746 & 0.632 & 0.540 & 0.453 & 3.25 \\ \rowcolor{Gray}
LTM-S (ours)     &  10\% ABC & 0.896 & 0.769 & 0.652 & 0.596 & 0.467 & 3.38 \\ \bottomrule
\end{tabular}
}
\vspace{-2mm}
\label{tbl:calvin}
\end{table}

\subsection{iThor Evaluation}

We next explore the ability to extend our method to benchmarks that involve ego motion of the robot (e.g. simple navigation tasks). Following prior work AVDC \citep{Ko2023LearningTA}, we evaluate on the iThor benchmark and present results in \Cref{app:ithor}. Results indite clear improvements of our proposed \modelname over naive baselines and prior work AVDC \citep{Ko2023LearningTA}.
The behaviour cloning (BC) baselines are implemented following \citet{Ko2023LearningTA}. 
Both AVDC and LTM-H (ours) are trained on the same data under common training settings for fair comparison. 

\begin{table}[ht]
\small
\centering
\caption{
    \textbf{Results on iThor Benchmark:} 
    We follow the iThor dataset based evalution setup used in AVDC \citep{Ko2023LearningTA} to demonstrate that our method generalizes to robot movement based control as well (i.e. where ego motion occurs). Results indicate that our method outperforms AVDC across categories and overall.
}
\label{app:ithor}
\vspace{-0.5em}
\def\arraystretch{1.3}  
\setlength\tabcolsep{1.0em}  
\scalebox{0.95}{
\begin{tabular}{lccccc}
\toprule
Method  & Kitchen & Living Room  & Bedroom & Bathroom & Overall \\ \midrule
BC-Scratch &  1.7 & 3.3  & 1.7  & 1.7  & 2.1  \\
BC-R3M     &  0.0 & 0.0  & 1.7  & 0.0  & 0.4  \\
AVDC       & 26.7 & 23.3 & 38.3 & 36.7 & 31.3 \\ \rowcolor{Gray}
LTM-H (ours) & 27.3 & 23.7 & 40.0 & 36.7 & 31.9 \\ \bottomrule
\end{tabular}}
\end{table}

Our \modelname was not explicitly designed for such ego-motion tasks, but nevertheless is capable of performing such tasks similar to AVDC. We take these results as a promising indication that \modelname can be further extended to better handle such ego-motion tasks.

\section{Relative Pixel Motion}
\label{app:relative_of}
A key design choice in our formulation is to represent pixel motion with respect to the current frame ($\vx_t$), rather than the previous frame ($\vx_{t+1}$) or some other frame. This aligns with the structure of our conditional diffusion model, which receives $\vx_t$ as a secondary conditioning input. Predicting the transformation from $\vx_t$ to the next frame allows the model to more directly focus on the visual cues present in the current state. In contrast, predicting motion from $\vx_{t-1}$ or some other different frame would require indirect reasoning over a non-visible state, introducing additional complexity. Hence our approach is to represent past pixel motion (e.g. $\vx_{t-1}$  to $\vx_{t}$) as $\vx_{t}$ to $\vx_{t-1}$ instead. While this may seem counterintuitive, we note how prior literature on image-pair-based optical flow prediction for video tasks has also found that defining motion in terms of a reference image—particularly the current frame that is visible—can lead to more stable and accurate flow estimates \citep{Liang2024MoVideoMV}. Moreover, our experiments representing previous motion in a different manner lead to subpar performance, standing as further evidence. 

We also experiment trying to predict an additional future motion relative to a future frame. We compare this against predicting that same future motion relative to the current frames. In this setting, the latter performs well while the former variant fails to learn meaningful motion signals predictions. 


\section{Language Embedding Model}
\label{app:lang_embed}
For the language embedding model, we employ the Universal Sentence Encoder (USE), a pre-trained model from \citep{Cer2018UniversalSE}. USE generates fixed-length vector representations of text, capturing rich semantic meaning, making it suitable for various natural language processing (NLP) tasks. Its widespread use in research, including works like OpenX \citep{padalkar2023open}, highlights its effectiveness in transforming textual input into meaningful embeddings even for robotic tasks. In our framework, the USE serves as a key component, encoding language instructions into dense vectors that are later used to guide the generation of motion representations. The model's ability to produce consistent and high-quality embeddings enables seamless integration between language and vision modalities, ensuring that our system can accurately interpret and respond to diverse language commands.

\section{Diffusion Model Details}
\label{app:dm_detail}

In our diffusion model training, input noising is applied by adding Gaussian noise to the target motion data (following standard settings from \citet{ho2020denoising}). The image condition input and the previous flow are not subject to this noising. 
The previous flow is corrupted with a 50\% chance. During corruption, a random amount of Gaussian noise is added. 
To ensure diverse and meaningful training, filtering and augmentation operations are performed on the frames as described next. 
The indices corresponding to consecutive frames ($i$ and $i+1$) are selected such that they maintain fixed intervals based on the video frame rate. Frames with zero optical flow (i.e., no motion) between $i$ and $i+1$ are filtered out to avoid irrelevant data. Additionally, to handle the completion of textual instructions, we introduce zero motion at the ends of videos, ensuring that these states map to a lack of motion when the instruction concludes. The visual inputs (images and optical flow) are cropped and resized, with appropriate transformations applied to the flow data to maintain consistency.

\section{Hand-Crafted Mapping Functions}
\label{app:handcraft_map}

\bhdr{Synthetic Environments:}
We follow the formulation of \citet{Ko2023LearningTA} using a segmentation map of robot controller and a depth map of environment. The generated pixel motions are converted into directions in 3D space to move the robot controller based on these dense maps. We direct the reader to \citet{Ko2023LearningTA} for further details. 

\begin{figure*}[t]
    \centering
    \begin{minipage}{\linewidth}
    {\small
        \hspace{1.5em} \textls[-60]{\texttt{Start State}} 
        \hspace{1.0em} \textls[-60]{\texttt{Predicted Motion Overlaid on Intermediate States}}
        \hspace{1.0em} \textls[-60]{\texttt{End State}}
    }
    \end{minipage}
    \begin{minipage}{0.03\linewidth}
    \centering
    \begin{tabular}{c}
         \rotatebox{90}{\texttt{Task-1}}   \\[2.7em] 
         \rotatebox{90}{\texttt{Task-2}}   \\[2.8em] 
         \rotatebox{90}{\texttt{Task-3}}   \\[2.9em] 
         \rotatebox{90}{\texttt{Task-4}}  \\
    \end{tabular}
    \end{minipage}
    \begin{minipage}{0.96\linewidth}
    \begin{minipage}{\linewidth}
        \centering\centering
        \includegraphics[width=\linewidth]{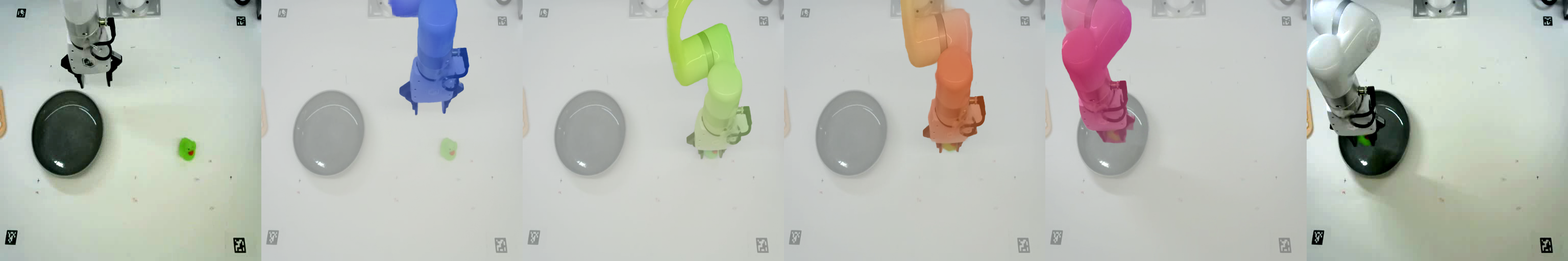}
    \end{minipage}
    \begin{minipage}{\linewidth}
        \centering\centering
        \includegraphics[width=\linewidth]{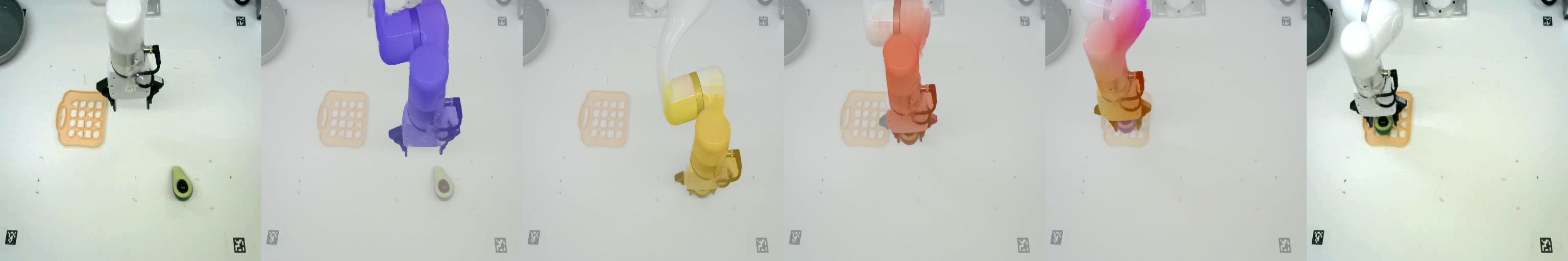}
    \end{minipage}
    \begin{minipage}{\linewidth}
        \centering\centering
        \includegraphics[width=\linewidth]{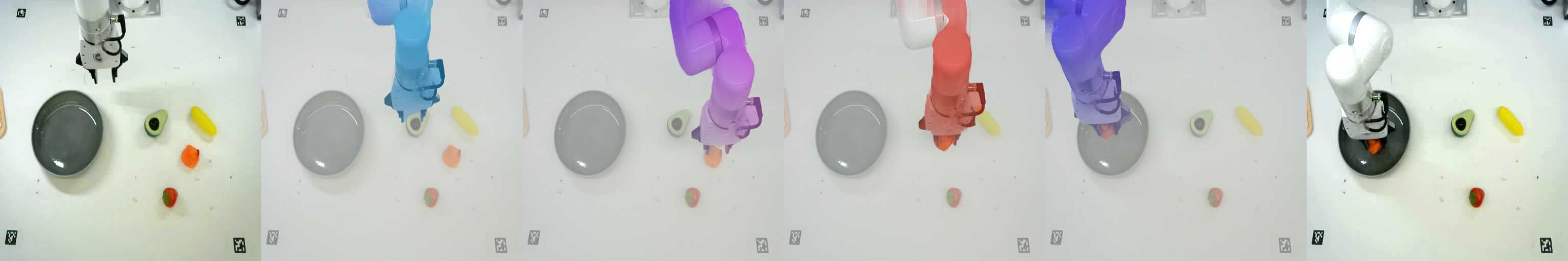}
    \end{minipage}
    \begin{minipage}{\linewidth}
        \centering\centering
        \includegraphics[width=\linewidth]{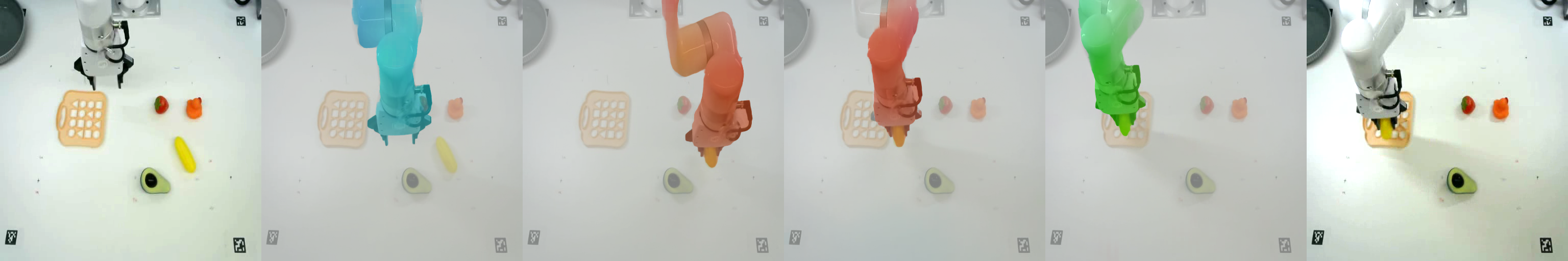}
    \end{minipage}
    \end{minipage}
    \vspace{-0.5em}
    \caption{
    \textbf{Real World Tasks:} 
    We illustrate the four real-world tasks following LLaRA~\cite{Li2024LLaRASR}. Start and end states are shown in the first and last columns, with predicted pixel motion (color indicates motion direction) overlaid on intermediate states. 
    \modelname performs these challenging tasks successfully (see results in \Cref{table:real}).
    }
    \label{fig:task_vis}
    \vspace{-0.5em}
\end{figure*}

\bhdr{Real World Environments:}
Motivated by \citet{Li2024LLaRASR}, we build our real world environment with a single plane assumption (e.g. table top manipulation) and map the predicted pixel motions for the robot controller center points onto the plane (using visual geometry). An initial camera calibration is performed for the environment to obtain necessary camera matrices. 
After extracting a start and end position for a manipulation task following this setting, our position to action vector conversion is identical to \citet{Li2024LLaRASR}.
Examples of these tasks are shown in \Cref{fig:task_vis}.

\section{Real World Experiments}
\label{app:real_world}

We perform four styles of real world experiments as illustrated in \Cref{fig:task_vis}. The language instructions for the four examples in this figure, where each belongs to one of the four task styles, are as follows: 
\begin{enumerate}[leftmargin=4em,noitemsep,topsep=0.0ex,itemsep=-0.5ex,partopsep=0ex,parsep=1ex]
  \item \texttt{Pick up the duck and place on the bowl.}
  \item \texttt{Pick up the duck and place on the tray.}
  \item \texttt{Pick up the avocado and place on the bowl.}
  \item \texttt{Pick up the corn and place on the tray.}
\end{enumerate}
Each task style contains similar textual commands that require some object manipulation in the table top environment. 
We select these following \citet{Li2024LLaRASR} to ensure fair comparisons to prior work. 

\section{Baseline Details}
\label{app:baselines}
Our key baselines are from AVDC \citep{Ko2023LearningTA} and LLaRA \citep{Li2024LLaRASR}. For both methods, we use their official implementations to replicate their results and evaluate ours under identical settings. For LLaRA, all results are reported on their inBC variant for fair comparison against our method (i.e. similar inputs during inference / no external scene object information). 
We also use official implementations for VPP \citep{Hu2024VideoPP}, Im2Flow2Act \citep{Xu2024FlowAT}, and ATM \citep{Wen2023AnypointTM} for evaluating those baselines. 
All these baselines are trained on the same data as our \modelname model.

\section{Detailed Ablations}
\label{app:ablate}

We discuss our ablations in \Cref{table:ablate} in detail in the following section.

\bhdr{System 2 Design Choices:}
We first ablate critical inputs to \textit{System 2 (Motion Generation)}.
Removing pretraining (``PT") leads to a modest performance drop (from 53.6\% to 53.1\%), indicating that while pretraining aids convergence, the framework remains effective with limited finetuning alone.
Removing the previous optical flow input (``Prev Flow") results in a larger decline to 50.2\%, validating the importance of temporal conditioning.
Ablating the language embedding leads to a significant drop (to 39.7\%), highlighting the necessity of semantic instruction guidance.
Finally, removing the visual input (``Img") results in near-random performance (5.6\%), confirming that visual grounding is essential.

\bhdr{High-Level Framework Design:}
We next evaluate several higher-level architectural decisions.
Removing the diffusion model (``No diffusion") and training a direct regression using an autoencoder based setup (using same architecture as our diffusion model but without noise inputs and with a single time-step for training and inference) leads to a sharp performance drop (to 16.2\%), underscoring the value of iterative, probabilistic modeling for motion generation.
Replacing input concatenation with cross-attention (``CA instead of concat") similarly degrades performance, suggesting that simple spatial concatenation is a more effective conditioning strategy for our setting.
Using a multi-action decoder within \textit{System 1} to run it at same frequency as our system 2 (``Sys-1 \& 2 same freq") results in slightly lower performance (48.7\%), indicating that our default action mapping is more effective.
Training only a learned \textit{System 1} without leveraging pixel motions generated by Sys-2 (``Only learned Sys-1") performs poorly (3.2\%), demonstrating that our System-1 module simply learns to map the generated pixel motion to robot manipulator actions.

\section{Pixel Motion Distribution Analysis}
\label{app:dist_analysis}

\begin{figure}[t]
\centering
\small    
\includegraphics[width=0.49\linewidth]{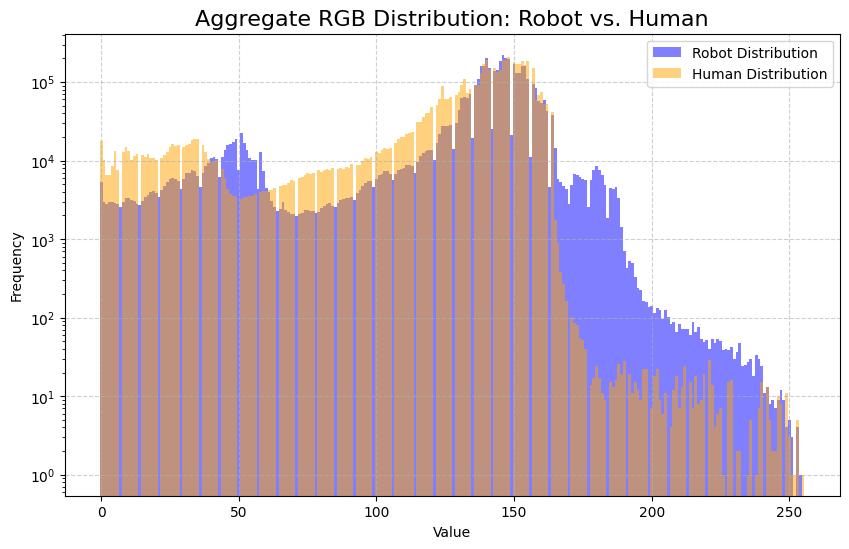}
\includegraphics[width=0.49\linewidth]{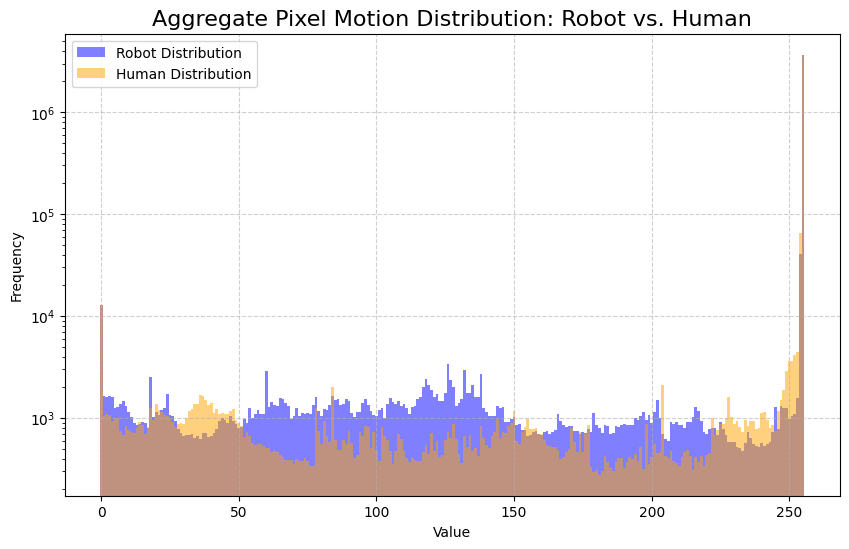}
\caption{
\textbf{Histogram Comparisons for RGB and Pixel Motion Distributions of Robot vs Human Demonstrations:}
We illustrate two histograms that compare the aggregate pixel value distributions of 40 human and 40 robot demonstrations. For the RGB distributions (left), the high degree of separation and low overlap between the distributions of the two groups indicate significant differences in appearance, resulting in a high Symmetrized KL Divergence of 0.7881. In contrast, the pixel motion distributions (right) show substantial overlap between RGB and pixel motion, demonstrating the strong similarity in kinematic patterns between human and robot demonstrations, leading to a much lower Symmetrized KL Divergence of 0.0199. These results suggest that pixel motion is a more embodiment-agnostic metric for comparing demonstrations.
}
\end{figure}

We present an analysis of the divergence between human and robot demonstration data using both RGB pixel values and pixel motion. To quantify the difference, we first aggregated the pixel value distributions from 40 human and 40 robot demonstrations, creating two distinct distributions for each data type. We then computed the Symmetrized Kullback-Leibler (KL) divergence to measure the difference between the human and robot distributions. The results, with a high KL divergence for RGB (0.7881) and a very low one for pixel motion (0.0199), indicate that the distributions of RGB pixel values are significantly different between human and robot demonstrations, while the distributions of pixel motion are remarkably similar. As illustrated in \Cref{app:dist_analysis}, this finding supports our hypothesis that pixel motion is a more embodiment-agnostic representation. The high divergence in RGB is a function of embodiment-specific factors such as lighting, skin tone, robot color, and background, which vary greatly between human and robotic forms. Conversely, the low divergence in pixel motion demonstrates that, regardless of the physical embodiment, the fundamental kinematic patterns of the motion itself are consistent, making it a more universal measure for learning from demonstrations.

\end{document}